\documentclass[10pt,twocolumn,letterpaper]{article}

\usepackage[pagenumbers]{cvpr} 

\usepackage{graphicx}
\usepackage{amsmath}
\usepackage{amssymb}
\usepackage{booktabs}
\usepackage{colortbl}
\usepackage{float}
\usepackage{multirow}
\usepackage{times}
\usepackage{epsfig}
\usepackage[accsupp]{axessibility}

\usepackage[pagebackref,breaklinks,colorlinks]{hyperref}

\usepackage[capitalize]{cleveref}
\crefname{section}{Sec.}{Secs.}
\Crefname{section}{Section}{Sections}
\Crefname{table}{Table}{Tables}
\crefname{table}{Tab.}{Tabs.}

\def\cvprPaperID{21} 
\def\confName{CVPR}
\def\confYear{2023}

\begin{document}

\title{Spectral Transfer Guided Active Domain Adaptation For Thermal Imagery}

\author{Berkcan Ustun\textsuperscript{1} \qquad\qquad Ahmet Kagan Kaya\textsuperscript{1,2} \qquad\qquad Ezgi Cakir Ayerden\textsuperscript{1,2} \quad\quad Fazil Altinel\textsuperscript{1}\\
\textsuperscript{1}Research Center, Aselsan Inc., Turkiye\\
\textsuperscript{2}Department of Electrical and Electronics Engineering, Middle East Technical University, Turkiye\\
{\tt\small \{berkcanustun, kagankaya, eayerden, faltinel\}@aselsan.com.tr}
}

\maketitle

\begin{abstract}
The exploitation of visible spectrum datasets has led deep networks to show remarkable success. However, real-world tasks include low-lighting conditions which arise performance bottlenecks for models trained on large-scale RGB image datasets. Thermal IR cameras are more robust against such conditions. Therefore, the usage of thermal imagery in real-world applications can be useful. Unsupervised domain adaptation (UDA) allows transferring information from a source domain to a fully unlabeled target domain. Despite substantial improvements in UDA, the performance gap between UDA and its supervised learning counterpart remains significant. By picking a small number of target samples to annotate and using them in training, active domain adaptation tries to mitigate this gap with minimum annotation expense. We propose an active domain adaptation method in order to examine the efficiency of combining the visible spectrum and thermal imagery modalities. When the domain gap is considerably large as in the visible-to-thermal task, we may conclude that the methods without explicit domain alignment cannot achieve their full potential.
To this end, we propose a spectral transfer guided active domain adaptation method to select the most informative unlabeled target samples while aligning source and target domains. We used the large-scale visible spectrum dataset MS-COCO as the source domain and the thermal dataset FLIR ADAS as the target domain to present the results of our method. Extensive experimental evaluation demonstrates that our proposed method outperforms the state-of-the-art active domain adaptation methods. The code and models are publicly available.\footnote[1]{\href{https://github.com/avaapm/STGADA}{https://github.com/avaapm/STGADA}}
\end{abstract}

\section{Introduction}
\label{sec:intro}

The latest state-of-the-art deep learning methods have led to a substantial improvement on computer vision and pattern recognition tasks such as classification and object detection with the use of RGB images \cite{rcnn, resnet, yolo, fasterrcnn}. Deep models trained on massive RGB datasets e.g., ImageNet \cite{imageNet}, MS-COCO \cite{mscoco}, Pascal-VOC \cite{pascalvoc}, etc. have shown considerable performance. Despite the progress of recent deep learning models trained on the visible spectrum images, low-lighting conditions prevent the most current models from performing competently on several real-world tasks. Since thermal cameras are more robust against these conditions, using them is advantageous for real-world applications such as military operations, security monitoring, autonomous driving, etc. However, large scale thermal datasets are not easily accessible to the public. Since the most advanced models require a massive amount of labeled data, models trained on thermal images struggle to perform as well as models trained on RGB images. One simple strategy to enhance the effectiveness of systems that use thermal imaging for classification and detection tasks is to take advantage of the additional information provided by visible spectrum images. Unfortunately, the models trained on visible spectrum datasets may perform poorly when the models tested on thermal images \cite{devaguptapu2019borrow, guan2019unsupervised, sgada2021}.

\begin{figure*}[ht]
   \centering
       \scalebox{0.50}{\includegraphics{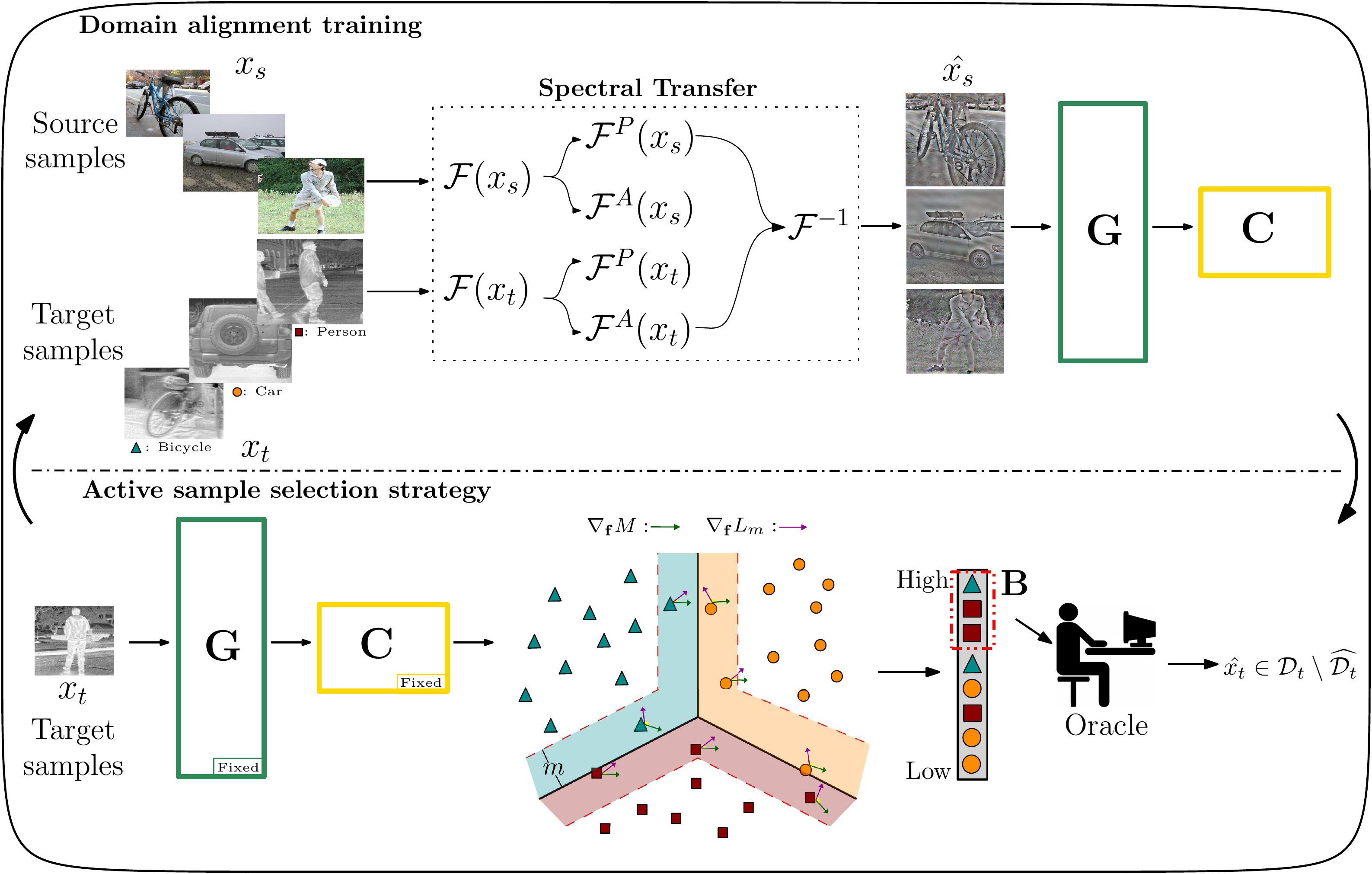}}
       \caption{An overview of our proposed spectral transfer guided active domain adaptation (STGADA) method. Source samples are subjected to spectral transfer with target samples. The amplitude component of the Fourier transform of the source samples $\mathcal{F}^A(x_s)$ are exchanged with that of target samples $\mathcal{F}^A(x_t)$. We  train our  feature extractor $\mathbf{G}$ and classifier $\mathbf{C}$ with the transformed source samples $x_{s\rightarrow t}$. Then, we proceed to active sample selection. We use the fixed feature extractor $\mathbf{G}$ and the classifier $\mathbf{C}$ to sample from the target dataset. In the feature space representation of the target samples, the query function measures how close a target sample to the decision boundary $M$. In addition, our sampling strategy considers how similar the direction of the gradient of the loss term $\nabla_{\mathbf{f}}L_m$ and that of margin term $\nabla_{\mathbf{f}}M$. Oracle annotates target samples of labeling budget \textbf{B} (dashed red box) to use in domain alignment training. \textit{Best viewed in color}.}
       \vspace{-3mm}
       \label{fig:overview}
\end{figure*}

In order to overcome the performance drop on target thermal domain while utilizing the source visible spectrum domain, the domain shift between the visible and thermal spectrum images must be taken care of. Domain adaptation (DA) methods intend to learn a mapping from both domains to the same feature space. Unsupervised domain adaptation (UDA) methods are based on the assumption that labels for target domain are not available \cite{dann,adda,mcdda}. Labeling all of the data is costly in terms of time and human resources. However, limitation in semantic annotations is more flexible in real world applications since a reasonable portion of the target samples can be annotated in consideration of the labeling costs and budget. In this sense, the concept of active learning is useful for determining which portion of the data should be annotated. Most active learning methods have recently concentrated on developing a query function that measures how representative and informative a sample is based primarily on uncertainty \cite{sinha2019variational}. However, domain shift between source and target samples hinders classical active learning query methods from selecting the samples which could increase the performance. The query methods that exploit the prediction of the task models trained only on source domain tend to sample some irrelevant target samples since most of the target data considered uncertain by the model. Therefore, active learning in consideration of domain shift problem is an important research area \cite{su2020active, fu2021transferable, xie2022learning}.

Recent state-of-the-art active domain adaptation methods consider the domain gap between the source and target datasets from two perspectives. Approaches such as \cite{rangwani2021s3vaada}, \cite{su2020active}, and \cite{xie2022active} focus on conducting an explicit domain adaptation to align the feature spaces of both domains. On the other hand, \cite{fu2021transferable} and \cite{xie2022learning} seek a solution to the problem of active domain adaptation without concerning explicit alignment of feature spaces. When the domain gap is considerably large as in the visible-to-thermal task, we may conclude that the methods without domain adaptation cannot achieve their full potential. In other words, selected target samples by the query functions become efficient in the case of a moderate domain shift. In that case, implementing additional domain adaptation algorithms on top of models does not yield a significant improvement. On the contrary, additional reduction of the domain gap may improve the performance in the RGB-to-thermal task since the domain gap is considerably large.

In this paper, we propose an active domain adaptation method to select the most informative samples while aligning visible and thermal domains as described in \cref{sec:method}. We employ Select-by-Distinctive-Margin (SDM) \cite{xie2022learning} as our base method. Based on our experimental evaluations in \cref{sec:exp}, SDM is more calibrated and achieves more successful results compared to the state-of-the-art active domain methods. Although SDM and other domain alignment-free active domain adaptation methods have achieved successful results, additional reduction of the domain gap may improve the performance in the RGB-to-thermal task since the domain gap is large compared to the domain gap in classical domain adaptation problems. In that regard, we apply a domain adaptation method to the SDM \cite{xie2022learning} to achieve the promised performance improvement.

Domain adaptation algorithms come as a module implemented additionally to a backbone which is used to extract features. Those modules are generally a few layered deep networks with complicated loss functions or regularization terms \cite{dann, cdan, mcdda}. Apart from popular domain adaptation approaches, recently, using frequency domain information in order to reduce the domain gap has become popular in terms of the simplicity of manipulating the frequency characteristics of the samples without increasing the complexity of the baseline model \cite{Li_2022, fda}. \cite{fda} proposes that interchanging the low frequency characteristics of the source and target samples reduces the domain gap. This method takes advantage of simplicity by applying the algorithm to the training samples just before they are input to the base model. Interchanging the low frequency components of source and target samples can be implemented as a transformation to the source samples that are used in the training of SDM \cite{xie2022learning}. With these in mind, we propose a spectral transfer guided active domain adaptation (STGADA) method in order to achieve higher performance in thermal domain via reducing the domain gap between visible and thermal IR spectra (\cref{fig:overview}). Spectral transfer is applied to the source domain samples by implementing the Fourier domain adaptation (FDA) \cite{fda} during training. The low-frequency components of the labeled source samples are changed with those of unlabeled target samples. This transfer reduces the domain gap between source and target domains. The transformed source samples are then fed to the backbone network. We sample from target thermal dataset according to a query function which uses the output of the fixed classifier i.e., the prediction scores of the overall model. This sampling process takes place in the specified epochs during training. Therefore, the training of the network continues after the sampling steps.  

During training and testing our approach, we exploit the large scale visible spectrum dataset MS-COCO \cite{mscoco} and thermal spectrum dataset FLIR ADAS \cite{flir}. Moreover, our method does not require paired samples of RGB and thermal images. We conduct extensive qualitative and quantitative analysis in order to evaluate our method. \cref{sec:exp} demonstrates that our approach outperforms the state-of-the-art methods with improving our base model. \cref{fig:reliability} depicts that given a model without domain alignment or active sampling i.e, the source only model, the calibration line is far from ideal. The union of predicted samples within a confidence interval does not yield proportionate accuracy. In contrast, our proposed model STGADA obtains more calibrated results leading to a narrower gap in the reliability diagram. Moreover, class imbalance is important to address since real world applications include imbalanced classes \cite{classimbalance}. Experimental studies in \cref{sec:exp} show that our model obtains more balanced performance than the state-of-the-art methods since those active domain adaptation methods tend to over-classify the majority category. 

Our contributions are summarized as follows:
\begin{itemize}
    \item We propose a simple yet efficient spectral transfer guided active domain adaptation approach for thermal IR spectrum. In order to efficiently reduce the domain gap between source and target domains, we employ a spectral transfer algorithm.
    \item We conduct extensive analysis in order to demonstrate the efficiency of our method in the domain adaptation setting where MS-COCO \cite{mscoco} is RGB source dataset and FLIR ADAS \cite{flir} is thermal target dataset. The results show that our method outperforms state-of-the-art active domain adaptation models by selecting more informative target thermal samples. 
\end{itemize}

\section{Related Work}
Deep learning methods have achieved promising results on computer vision tasks by using RGB images. However, real-world conditions such as low-lighting challenge these models. Therefore, recent studies \cite{sgada2021, guo2019domain, kaist, BMVC2016_73, saponaro2015material, Uzun_2022_CVPR} investigated performance on the classification and object detection tasks in relation to the consequences of combining visible spectrum and thermal image modalities. Motivated by these studies, in this work, we take advantage of supplementary information provided by visible spectrum images to enhance classification performance on thermal imagery without requiring RGB-to-thermal image pairings since large-scale thermal datasets are not publicly available. To evaluate the effectiveness of combining RGB and thermal images, we propose an active domain adaption approach. 

Active learning algorithms choose the samples to annotate with the semantic annotations rather than having access to the labels beforehand \cite{settles2011theories}. In order to choose the most disagreeing samples, query by committee approaches exploited many classifiers and measure their discrepancy on the task on unlabeled data using methods like low-dimension projections \cite{gilad2006query}. Query by uncertainty methods employed classification margin as a selection criteria \cite{balcan2007margin}, multi-class classification margin\cite{MCALforclassication}, learning loss \cite{yoo2019learning}, number of false-positive or false-negative pixels in images \cite{aghdam2019active}, discrepancy between two auxiliary classifiers \cite{Cho2021MCDALMC}. To choose a collection of unique samples, representativeness approaches often utilized clustering or core-set selection \cite{ALforCNN, sinha2019variational, agarwal2020contextual}. Although, these methods achieved successful results on active learning problem, none of them considers the potential domain shift between labeled and unlabeled data.  

Domain adaptation, in other words, the transfer of models trained on labeled source domains to unlabeled or partially labeled target domains, has drawn a lot of interest \cite{long2013transfer, dann, bousmalis2016domain, das2018unsupervised}. Ganin et al. \cite{dann} proposed a gradient reversal layer to align source and target domains using the feature encoder which is trained to maximize the loss of the domain discriminator while minimizing the loss of the classifier. Saito et al. \cite{mcdda} proposed to reduce the gap by employing adversarial learning with two task-specific classifiers and a feature extractor to align source and target domains. Besides from popular adversarial domain adaptation methods, using frequency information has gained interest to achieve distribution alignment. Yang et al. \cite{fda} proposed Frequency Domain Adaptation (FDA) method to overcome domain shift problem by simply replacing the low frequency component of a source sample with a target sample since the low frequency component of the samples provides the domain invariant properties. Therefore \cite{fda} introduces Frequency Domain Adaptation (FDA). Recently, Akkaya et al. \cite{sgada2021} proposed a pseudo-labeling guided unsupervised domain adaptation method by using large-scale RGB dataset MS-COCO as source domain and thermal dataset FLIR ADAS as target domain. Although adversarial and frequency-level domain adaptation methods have achieved successful experimental results, active domain adaptation can boost the performance of domain adaptation by labeling small portion of unlabeled target dataset. Since a reasonable portion of the target samples can be annotated in consideration of the labeling costs and budget, domain adaptation can be combined with active learning to improve performance on RGB-to-thermal domain adaptation problem.

Active domain adaptation has gained increasing attention in recent years due to its potential to improve model performance in real-world scenarios. Su et al. \cite{su2020active} proposed active adversarial domain adaptation (AADA) method to employ adversarial training of a feature generator and a domain discriminator for domain alignment and present a sample selection approach based on variety and uncertainty of target samples. Using a new sample selection methodology known as CLUE, Prabhu et al. \cite{prabhu2021active} suggested clustering target data embeddings weighted by uncertainty and choosing nearest neighbors to the cluster centroids for annotation while aligning domains using \cite{dann}. Fu et al. \cite{fu2021transferable} proposed transferable query selection (TQS) method in their sample selection strategy which consists of three factors: discrepancy among a group of five classifiers, margin of projected class probabilities for a target sample, and sample diversity from the source domain via a domain discriminator. To choose the most informative subset from the target domain to annotate, Rangwani et al. \cite{rangwani2021s3vaada} proposed S$^3$VAADA, which provides a submodular set-based information criterion that consists of three scores: virtual adversarial pairwise score, diversity score, and representativeness score. Based on energy-based models, Xie et al. \cite{xie2022active} find target samples which are particularly distinctive to the target distribution by taking use of free energy biases. More recently, Xie et al. proposed Select-by-Distinctive-Margin (SDM) method which comprises of a maximum margin loss and a margin sampling algorithm. Using greatest margin loss, SDM chooses target samples based on their relationship to a few hard instances from the source domain which are near to the decision boundary between source and target domain. Although SDM presents remarkable results, there is still room for improving performance on the tasks which have large domain gap by applying domain alignment.

Our approach differs from the past approaches that it manages the domain shift reduction, which is crucial in the case of the transfer of information from visible to thermal spectrum, without being overly complicated. To the best of our knowledge, there is no active domain adaptation study in the literature of thermal image classification. While the SDM \cite{xie2022learning} employs no extra classifiers for query by committee or complex architectures for aligning the distribution for the source and target domains, this lack of alignment leaves a space for improvement by reducing the domain gap. In addition, our choice of FDA \cite{fda} introduces no extra learnable parameters although it shows proper capability on dealing with the domain shift problem. 

\section{Proposed Method}
\label{sec:method}

Our proposed spectral transfer guided active domain adaptation method is illustrated in \cref{fig:overview}. 

First, we change low-frequency components of the labeled source samples with those of unlabeled target samples. We take Fourier transform $\mathcal{F}$ of source and target samples. The output of the Fourier transform divides into two; phase and amplitude components. Then, the amplitude component of the target sample is multiplied with a mask $M_\beta$ whereas that of the source sample is multiplied with the complement of that mask $(1-M_\beta)$. These amplitude components are summed to generate the new amplitude component. With this process, we control the portion size of the low frequency part that is exchanged. The phase component of the source sample and the newly generated mixed amplitude component are subjected to inverse Fourier transform $\mathcal{F}^{-1}$. In this way, a spectral transfer which aids to reduce the domain gap between two datasets occurs between source and target samples. The transformed source samples are then fed to the feature extractor $ \mathbf{G} $. A linear classifier $\mathbf{C}$ is trained along with the $ \mathbf{G} $. 

We sample from the target dataset according to a query function $Q(\cdot)$ which uses the output of the fixed $ \mathbf{C} $ in other words, the prediction scores of the overall model. The query function $Q(\cdot)$ also uses the backward gradient of the loss function of the backbone network. Since target samples have no labels, this gradient is estimated for target samples. Once the selected target samples are labeled by oracle, we utilize the selected target samples in training of $ \mathbf{G} $ and $\mathbf{C}$ with their labels. This sampling process takes place in the specified epochs during training. Therefore, the training of the network continues after the sampling steps.

In the inference, the target samples pass through the backbone network which consists of the feature extractor $ \mathbf{G} $ and the classifier $\mathbf{C}$. Thus, inference with our proposed approach is remarkably simple.   

A general definition of active domain adaptation, spectral transfer, and spectral transfer guided active domain adaptation procedures of our proposed approach are described in \cref{ada}, \cref{st}, and \cref{stgada}, respectively. 

\subsection{Active Domain Adaptation}
\label{ada}
In active domain adaptation, the source domain has $ n_s $ labeled samples which is represented as $\mathcal{D}_s =  \{(x^i_s, y^i_s)\}_{i=1}^{n_s}$ with samples $x^i_s$ and their semantic annotations $y^i_s \in \{ 1,2,\dots K \}$ where $K$ is the number of classes. The target domain has $n_t$ unlabeled samples which is denoted as $ \mathcal{D}_t = \{(x^j_t)\}_{j=1}^{n_t} $. In addition, we describe a labeled target set which is initially an empty set $\varnothing$, as $\widehat{\mathcal{D}_t}$. The unlabeled target data is sampled several times. The annotators assign their labels to the selected target data, $\hat{x_t} \in \mathcal{D}_t \setminus \widehat{\mathcal{D}_t} $. The model can be trained with the new labeled set, $\widehat{\mathcal{D}_t} \cup \mathcal{D}_s$, and then it will be exploited to sample new unlabeled data from the set $\mathcal{D}_t \setminus \widehat{\mathcal{D}_t}$. Given a budget $\textbf{B}$, this procedure continues until the annotated number of target samples achieves the predefined budget $|\widehat{\mathcal{D}_t}| = \textbf{B}$. 

\subsection{Spectral Transfer}
\label{st}

In this subsection, we elaborate the spectral transfer procedure exploited in our proposed method (dashed box in the upper part of \cref{fig:overview}).

For the Fourier transform $\mathcal{F}$ of an RGB image, $\mathcal{F}^A, \mathcal{F}^P: \mathbb{R}^{H \times W \times 3} \rightarrow \mathbb{R}^{H \times W \times 3}$ represent the amplitude and phase components, respectively. For simplicity of explanation, Fourier transform $\mathcal{F}$ of a single channel image $x$ is:
\begin{equation}
    \mathcal{F}(x)(u,v) = \sum_{h=0}^{H-1}\sum_{w=0}^{W-1} x(h,w)e^{-j2\pi(\frac{h}{H}u + \frac{w}{W}v)}.
\end{equation}
\vspace{-3mm}

This can be efficiently implemented by FFT algorithm proposed in \cite{fft}.
In order to map frequency domain signals back to image space, we perform inverse Fourier transform $\mathcal{F}^{-1}$. A mask $M_\beta$ is defined whose value is zero except for the region in the center $(h,w) \in [- \beta H: \beta H,-\beta W:\beta W]$ where $\beta \in (0,1)$ with the assumption that the center is $(0,0)$. We can define the spectral transfer from $x_t \in \mathcal{D}_t$ to $x_s \in \mathcal{D}_s$
as:
\vspace{-3mm}
\begin{equation}
    \hat{x_{s}} = \mathcal{F}^{-1}([M_\beta \circ \mathcal{F}^A(x_t)+(1-M_\beta) \circ \mathcal{F}^A(x_s), \mathcal{F}^P(x_s)])
\end{equation}
where the low-frequency component of the amplitude of the source sample is replaced by that of the target sample. The operation $\circ$ denotes the point-wise multiplication of the image and the mask matrices. The phase component of the source image is unchanged. This altered spectral signal is then mapped back to the image domain. The image $\hat{x_{s}}$ has the same content as $x_s$ while it constitutes the style and appearance of the sample $x_t$ from $\mathcal{D}_t$. 

\subsection{Spectral Transfer Guided Active Domain Adaptation}
\label{stgada}

In the task of active domain adaptation, all labeled data $\mathcal{D}_s \cup \widehat{\mathcal{D}_t}$ is used. This situation arises the problem of bias towards areas in the source domain where data samples are densely placed. Class imbalance problem in real-world applications amplifies this bias. Therefore, a network which is trained using all of the training data may be overfit to the abundant class. Moreover, this bias prevents the $Q(\cdot)$ from selecting informative target samples. Arising from this fact, a categorical-wise margin loss with its selective property can be employed to supervise the network. The proportional contribution of the source samples to the backward gradient concerning their margin size helps the network to focus on samples that have different difficulties. The consideration of uncertain source samples helps network to prevent the overfitting to the abundant class. Furthermore, we employ a max-logit regularizer to ensure that the network always assigns large scores to the prediction on ground truth class. Assigning large scores to the prediction on the ground truth class is important for calibration. We expect that the accuracy of the union of samples within a confidence level is similar with that confidence for a calibrated model. Concerning calibration, confidence score needs to be as large as accuracy. Hence, considering the spectral transfer in \cref{st} and adaptive margin loss, the overall loss function can be denoted as:
\vspace{-2mm}
\begin{align}
\label{loss}
    L_m(\hat{x_{s}}, y_s)  & = \sum_{i\neq j}[\gamma_i[m - \mathbf{C}(\mathbf{G}(\hat{x_{s}}))_j+ \mathbf{C}(\mathbf{G}(\hat{x_{s}}))_i]_+ \nonumber \\
    &- \mathbf{C}(\mathbf{G}(\hat{x_{s}}))_j] \\        
    \gamma_i & = 1-\frac{\mathbf{C}(\mathbf{G}(\hat{x_{s}}))_j -  \mathbf{C}(\mathbf{G}(\hat{x_{s}}))_i}{m}\nonumber    
\end{align}
where $[x]_+$ is the operation $max(0, x)$, the subscripts $i$ and $j$ denotes the $i$-th and $j$-th entry of vectors, respectively. $m$ is a hyper-parameter to control the margin width. Here, only the samples which have a confidence score of ground-truth class close to the classification scores of other classes can contribute to the backward gradient. In addition, we control the contribution by using $\gamma_i$ which is proportional to the similarity between scores of the ground truth class and those of other classes. 

While this loss function enhances the spaces between different categorical clusters, we can focus on the target samples that are placed near smaller gaps between categorical clusters in the feature space. Furthermore, ensuring the gradient from loss and the margin sampling term present similar orientations in the feature space will guarantee fast convergence to a robust state. Let the margin term be defined as:
\begin{equation}
    M(x_t) = (1 - (\mathbf{p}_{1^*}- \mathbf{p}_{2^*}))
\end{equation}
where;
\begin{equation}
\label{p}
   \mathbf{p} = \textbf{softmax}( \mathbf{C}(\mathbf{G}(x_{t})))
\end{equation}
and subscript $1^*$ and $2^*$ denotes the maximum and second maximum prediction scores.

By considering these, the overall query function for a sample $x$ can be denoted as:
\vspace{-3mm}
\begin{equation}
\label{query}
    Q(x) = M(x) + \lambda\left<\nabla_{\mathbf{f}}{L_m(x, y)}, \nabla_{\mathbf{f}}{M(x)}\right>
\end{equation}
where $\mathbf{f} = \mathbf{G}(x)$. In \cref{query}, $\lambda$ is a hyper-parameter and operation $\left<\cdot,\cdot\right>$ defines the Cosine Similarity.
Since we do not have access to the label of the target samples, the gradient from the loss term cannot be calculated, but can be estimated:
\begin{equation}
    \nabla_{\mathbf{f}}{\hat{L}_m(x, y)}=\mathbf{p}_{1^*}\nabla_{\mathbf{f}}{L_m(x, 1^*)}+\mathbf{p}_{2^*}\nabla_{\mathbf{f}}{L_m(x, 2^*)}.
\end{equation}

With the query function $Q(x)$, we can select the most informative target samples that are placed near confused decision boundaries and ensure fast convergence. Similar with the categorical margin loss in \cref{loss}, the query function $Q(x)$ considers the uncertain samples. 

\begin{figure*}[ht]
   \centering
         \scalebox{0.85}{\includegraphics[trim={1.5cm 1cm 1cm 0cm}, clip, width=\textwidth]{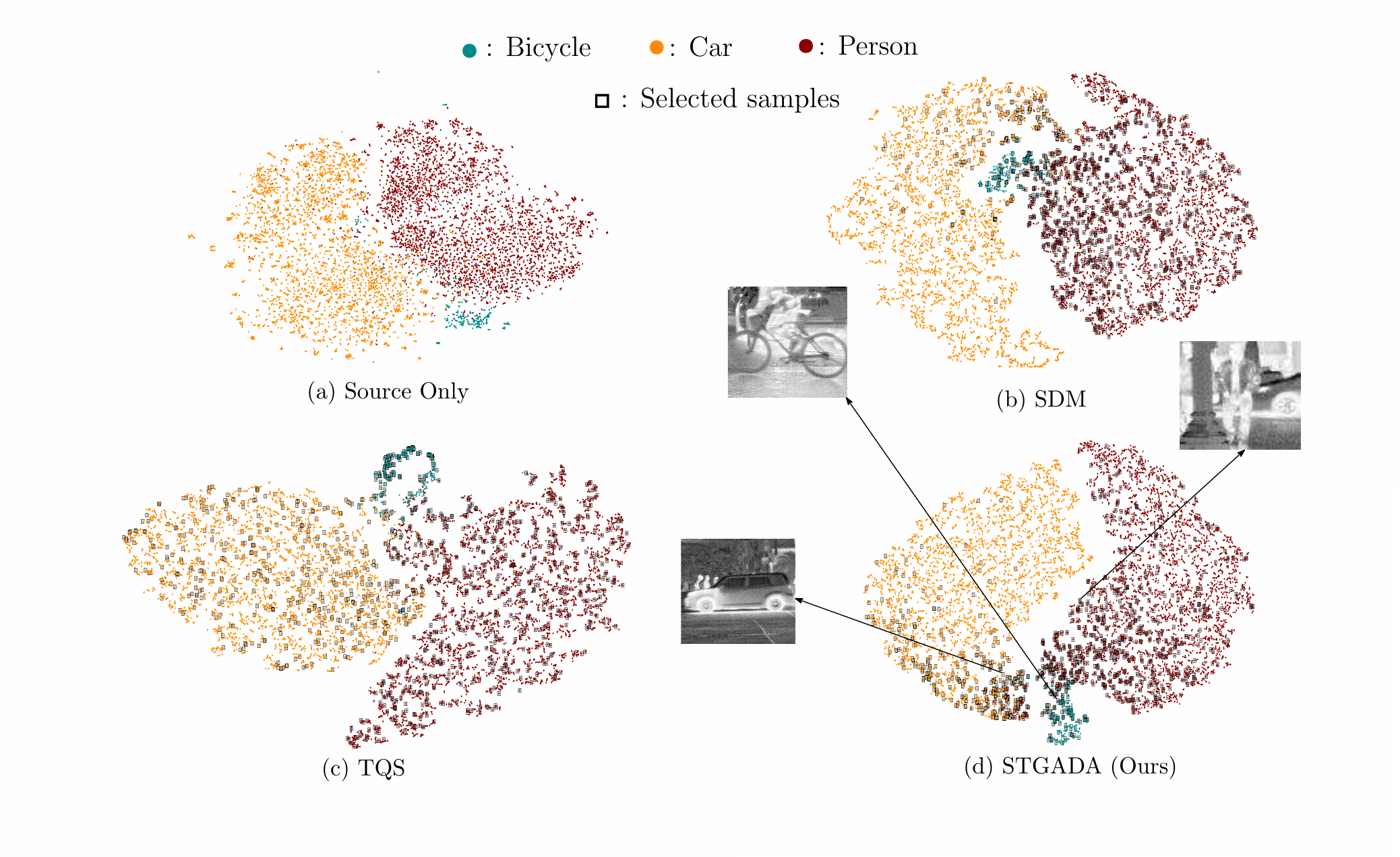}}
    \caption{The t-SNE \cite{tsne} visualization of network activations on target thermal domain generated by source only model (a), our base method SDM \cite{xie2022learning} (b), TQS \cite{fu2021transferable} (c), and our proposed method STGADA (d). We visualize one sample per each class selected by our proposed method STGADA. \textit{Best viewed in color}.}
    \vspace{-5mm}
    \label{fig:tsne}
\end{figure*}
\vspace{-3mm}
\section{Experiments}
\label{sec:exp}

We carry out comprehensive analyses and compare our proposed method with a number of state-of-the-art unsupervised domain adaption techniques.

\subsection{Datasets}
\vspace{-1mm}
We employed the RGB-to-thermal domain adaptation setting which is recently proposed by Akkaya et al. \cite{sgada2021}. The setting consists of two datasets, namely, FLIR ADAS \cite{flir} as a thermal dataset and MS-COCO \cite{mscoco} as an RGB dataset. 

FLIR ADAS \cite{flir} contains both RGB images and thermal images. We utilized only the thermal images in our experiments. As in \cite{sgada2021}, we used three classes from FLIR ADAS dataset: bicycle, car, and person. MS-COCO \cite{mscoco} is a publicly available large-scale visible spectrum dataset that contains the same classes as FLIR ADAS \cite{flir}. Therefore, MS-COCO \cite{mscoco} is considered visible spectrum dataset in our experiments in parallel with \cite{sgada2021}.

The images in both datasets originate from the square bounding box annotations of the objects i.e, bicycle, car, person. The extracted object images are resized to $224 \times 224$. Consequently, the thermal dataset includes 4,137 samples of bicycles, 43,734 samples of cars, and 26,294 samples of person images. The RGB dataset consists of 5,732 samples of bicycles, 38,453 samples of cars, and 209,162 samples of person images.  

\begin{table}[ht]
\renewcommand{\arraystretch}{1.3}
\caption{Per-class classification performance comparison.}
\label{table:results}
\vspace{-3mm}
\centering
\scalebox{1.07}{
\begin{tabular}{@{}l c c c >{\columncolor[gray]{0.8}}c}
\toprule
    Method & \rotatebox{90}{Bicycle } & \rotatebox{90}{Car } & \rotatebox{90}{Person } & \rotatebox{90}{\textbf{Average }}\\
\hline
Source only & 69.89 & 83.89 & 86.52 & 80.10 \\
Random & 63.91 & 96.73 & 97.87 & 86.17 \\
Entropy \cite{wang2014new} & 72.41 & 97.38 & 96.49 & 88.76 \\
\hline
S$^3$VAADA \cite{rangwani2021s3vaada} & 56.78 & 72.13 & \textbf{98.77} & 75.89 \\
EADA \cite{xie2022active} & 74.18 & 94.41 & 96.52 & 88.37 \\
AADA \cite{su2020active} & 81.15 & 97.81 & 94.65 & 91.20  \\
CLUE \cite{prabhu2021active}& 86.21 & 96.95 & 91.75 & 92.30 \\
SDM \cite{xie2022learning}& 88.05 & 98.35 & 96.51 & 94.30 \\
TQS \cite{fu2021transferable}& 90.34 & 97.93 & 96.41 & 94.89 \\
STGADA (Ours) & \textbf{90.57} & \textbf{98.96} & 96.53 & \textbf{95.35} \\
\bottomrule
\end{tabular}}
\vspace{-3mm}
\end{table}
\vspace{-1mm}
\subsection{Implementation Details}

We followed the training procedure of \cite{xie2022learning} in our experiments. We used ResNet-50 pre-trained on ImageNet feature extractor $\mathbf{G}$ for all approaches. The FDA \cite{fda} method is implemented on top of our backbone as a spectral transformation. During the training of our method, we set the batch size to 32. The number of epochs was determined to be 50. AdaDelta optimization algorithm is employed to update the parameters. The learning rate was set to 0.5. The  parameters $\lambda$ and $\beta$ were chosen as 0.001 and 0.033, respectively. The margin size $m$ was selected as 1. Throughout the experiments, we perform 5 rounds of active sampling. 2\% of target samples are labeled at every round. Following the standard active domain adaptation approaches \cite{su2020active, fu2021transferable, xie2022learning}, we simulate oracle annotations by using the ground truth. Labeled target data are extracted from unlabeled target set and added to source set. For the first 10 epochs, the training proceeded with the initial source data. In epochs 10, 12, 14, 16, and 18, we perform our sample selection strategy. 

We implemented our proposed method using PyTorch framework \cite{pytorch}. Implementation details, models, and the code were made publicly available at \href{https://github.com/avaapm/STGADA}{https://github.com/avaapm/STGADA}.
\vspace{-3mm}
\subsection{Results}
\vspace{-1mm}
In the experiments, the visible spectrum is chosen as the source domain, and the thermal IR spectrum is chosen as the target domain. As a general practice in domain adaptation, we denote \textit{source only} as the target dataset performance of a model trained with only the source dataset. Performance of \textit{source only} model serve as baseline for the lower bound performance.

\begin{figure*}[ht]
\centering
   \begin{subfigure}[b]{0.33\textwidth}
        \includegraphics[width=\textwidth]{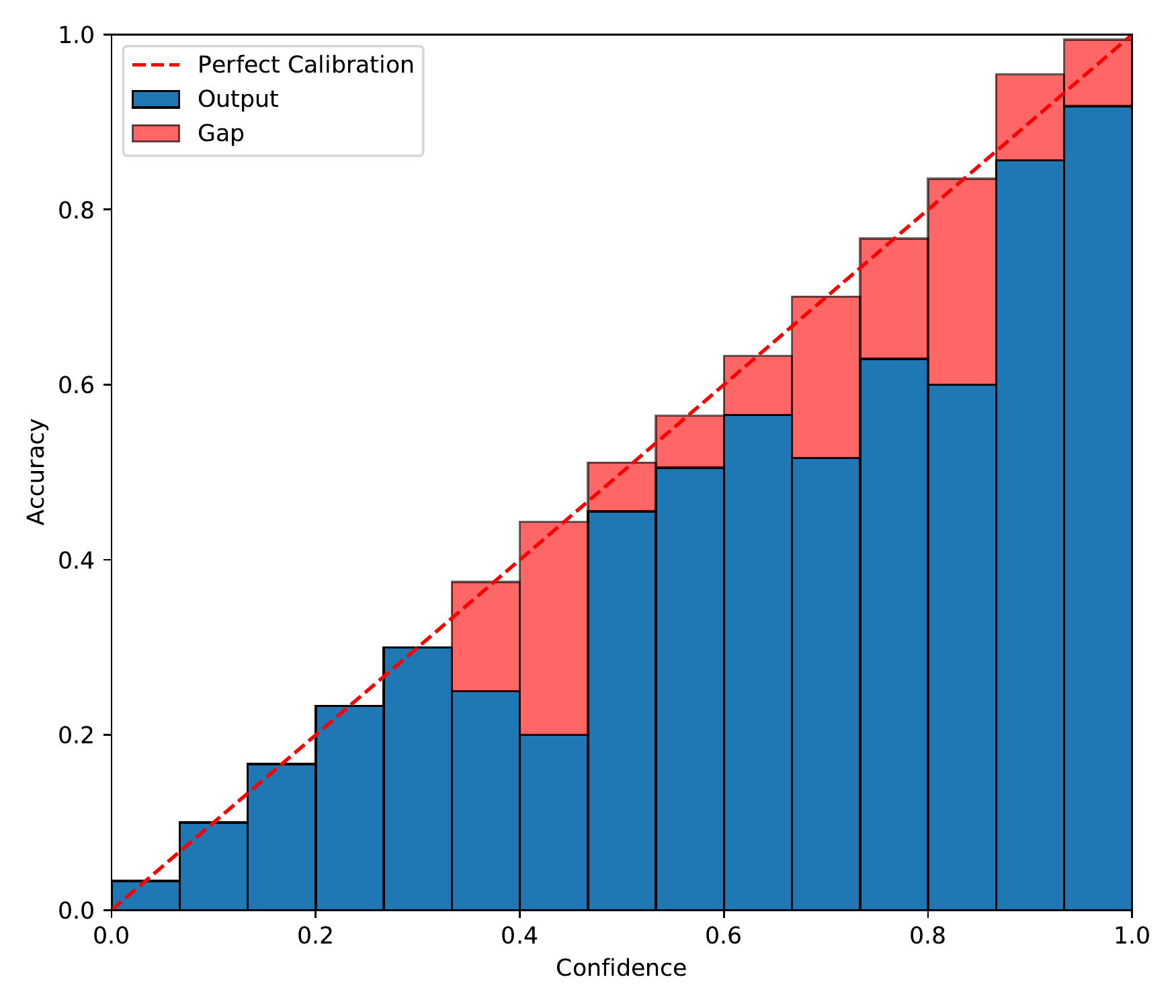}
        \caption{Source only}
         \label{fig:so_rel}
   \end{subfigure}
   \hspace{12mm}
   \begin{subfigure}[b]{0.33\textwidth}
        \includegraphics[width=\textwidth]{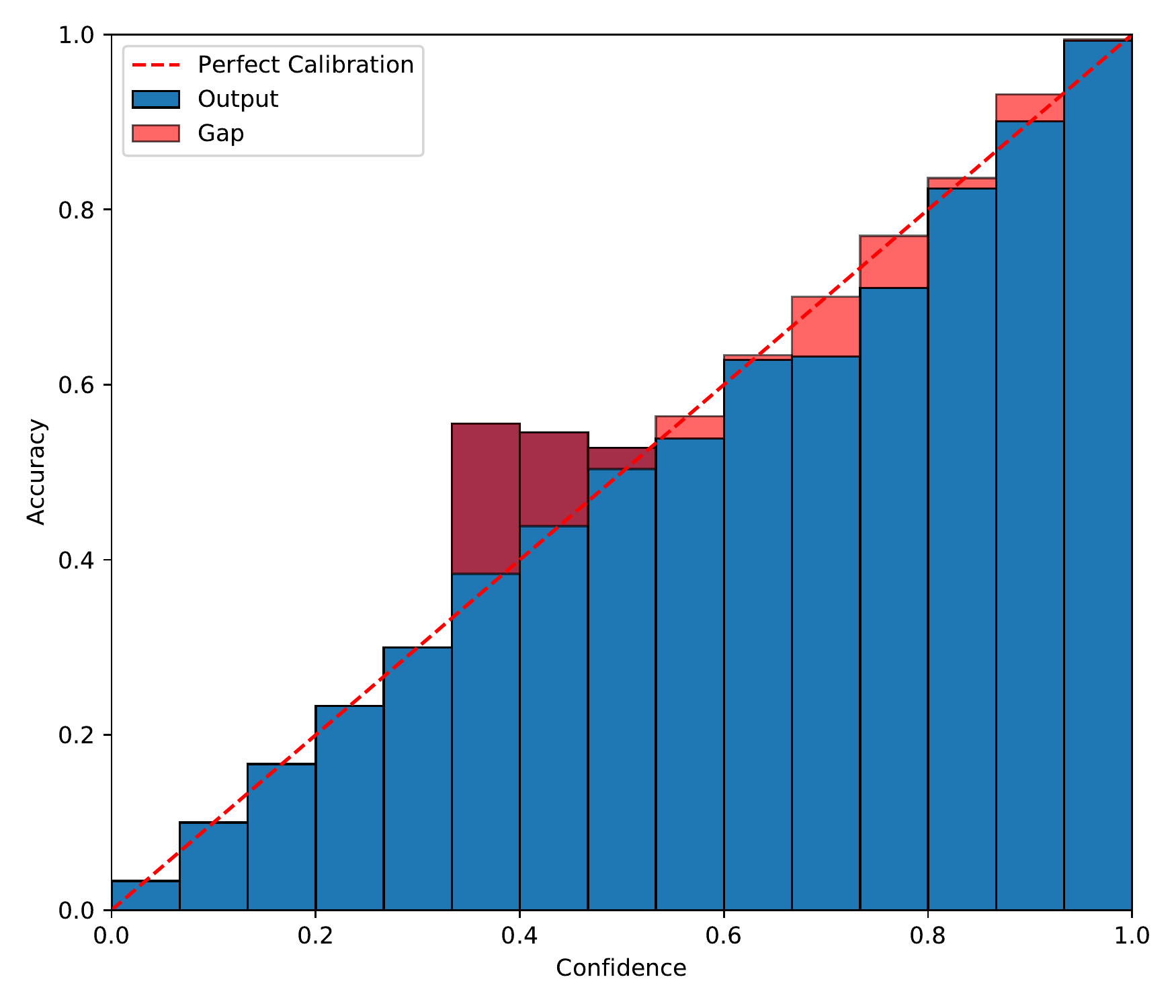}
        \caption{SDM}
         \label{fig:sdm_rel}
   \end{subfigure}
      \begin{subfigure}[b]{0.33\textwidth}
        \includegraphics[width=\textwidth]{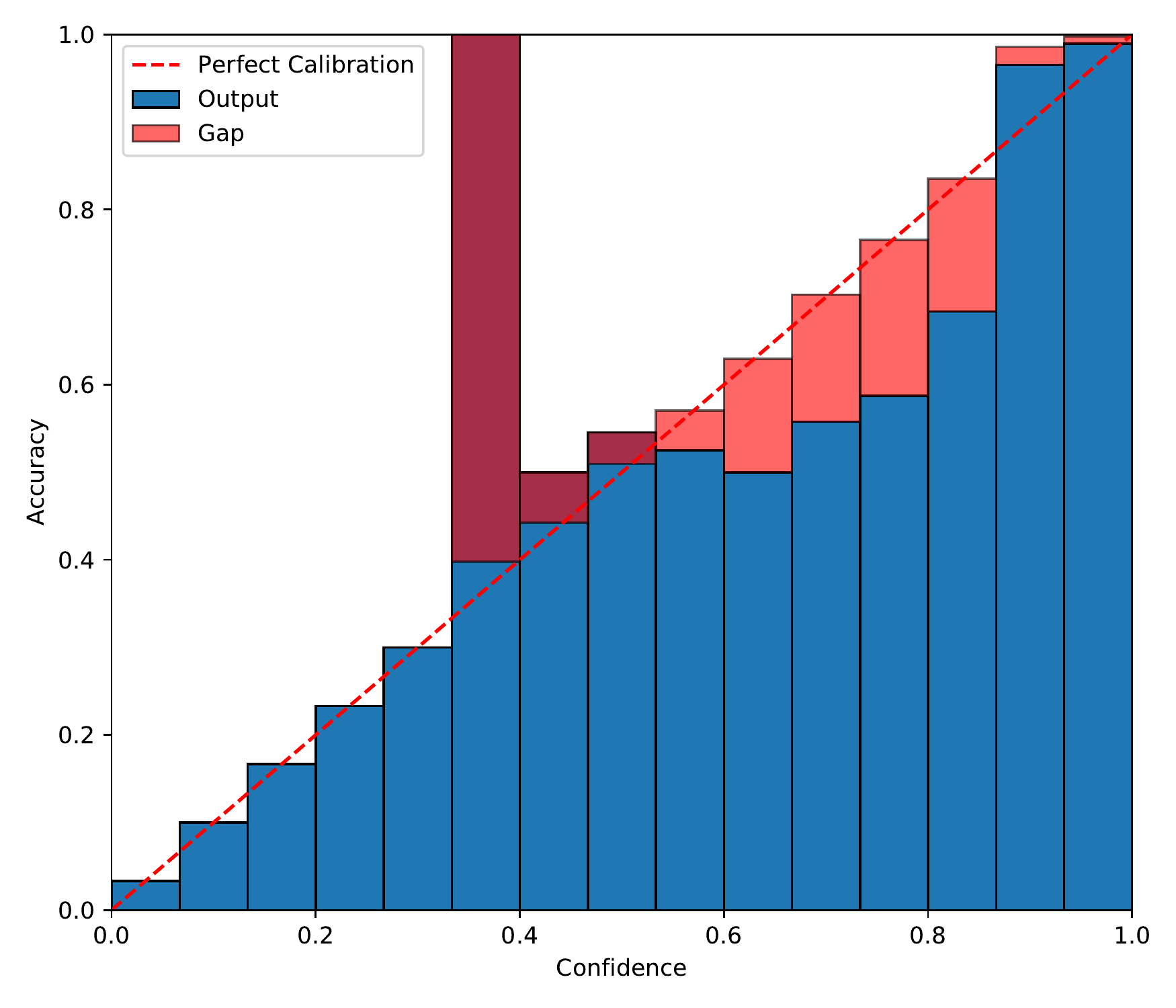}  
        \caption{TQS}
         \label{fig:tqs_rel}
   \end{subfigure}
   \hspace{12mm}
      \begin{subfigure}[b]{0.33\textwidth}
        \includegraphics[width=\textwidth]{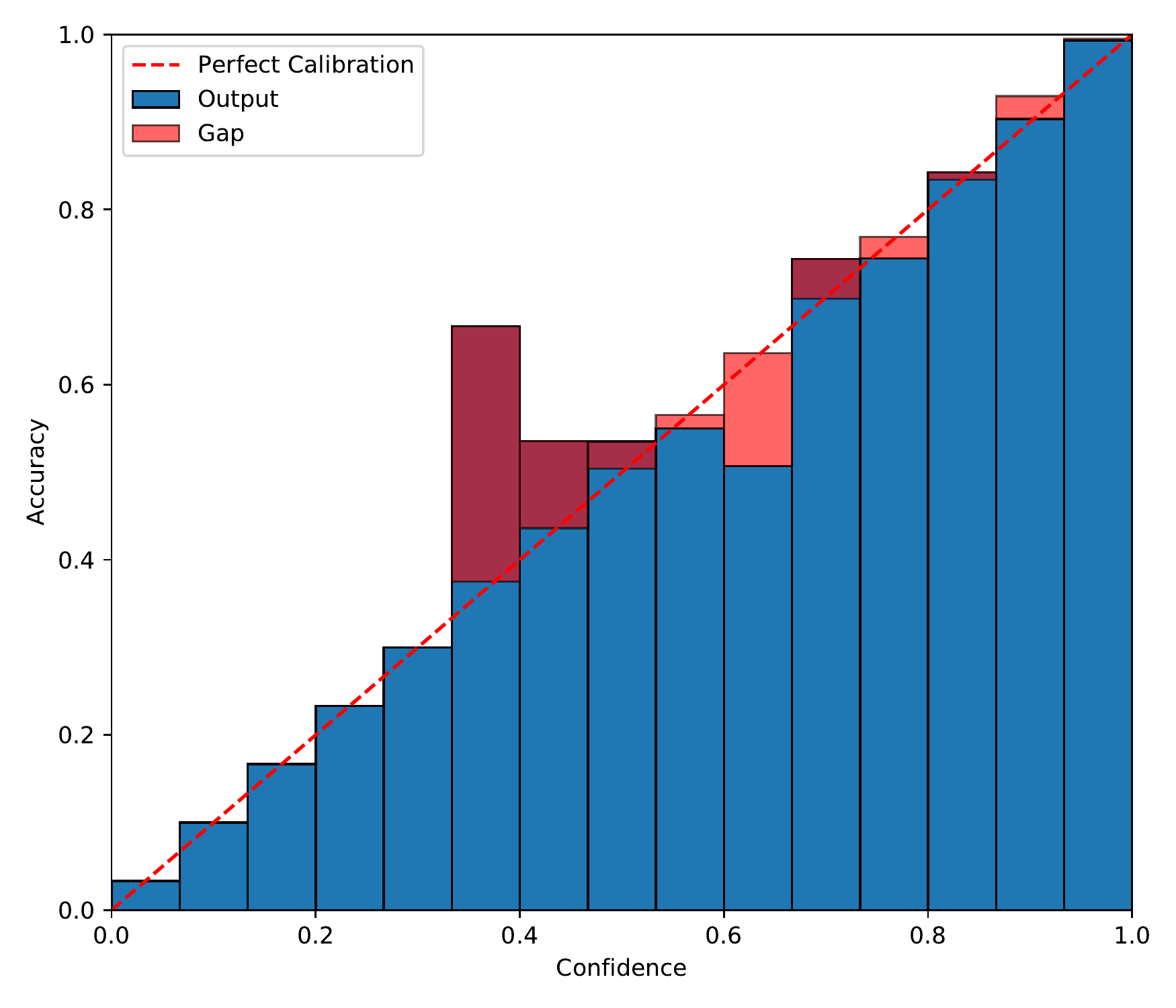}   
        \caption{Our method}
         \label{fig:stgada_rel}
   \end{subfigure} 
   \vspace{-3mm}
   \caption{Visualization of reliability diagrams \cite{ece} of source only model (a), SDM \cite{xie2022learning} (b), TQS \cite{fu2021transferable} (c), and our proposed method STGADA  (d). Given target domain images, our proposed method STGADA obtains more calibrated results compared to the source only model and TQS \cite{fu2021transferable} while showing comparable calibration gap with SDM \cite{xie2022learning}.}
   \vspace{-5mm}
   \label{fig:reliability}
\end{figure*}

\textbf{Quantitative Analysis.} We compare our proposed method STGADA with several state-of-the-art active domain adaptation methods, namely Active Adversarial Domain Adaptation (AADA) \cite{su2020active}, Clustering Uncertainty-weighted Embeddings (CLUE) \cite{prabhu2021active}, Energy-based Active Domain Adaptation (EADA) \cite{xie2022active}, Submodular Subset Selection for Virtual Adversarial Active Domain Adaptation (S$^3$VAADA) \cite{rangwani2021s3vaada}, Transferable Query Selection (TQS) \cite{fu2021transferable}, and Select by Distinctive Margin (SDM) \cite{xie2022learning}. Moreover, we compare our proposed method with baseline active learning sampling strategies: random sampling and entropy sampling \cite{wang2014new}. There are no published findings for our dataset in these studies since these approaches do not take domain adaptation into account for thermal datasets. As a result, we trained and evaluated each of these approaches using our dataset. We also conduct an analysis on the calibration of the source only model, SDM \cite{xie2022learning}, TQS \cite{fu2021transferable} and our proposed model STGADA. The Expected Calibration Error (ECE) \cite{ece} values are reported in \cref{table:ece} and \cref{fig:reliability} depicts the reliability diagram of the models.  

\begin{table}[ht]
\renewcommand{\arraystretch}{1.3}
\caption{Expected Calibration Error (ECE) \cite{ece} results. The lower the better.}
\label{table:ece}
\vspace{-3mm}
\centering
\scalebox{1.07}{
\begin{tabular}{@{}l c c }
\toprule
    Method & ECE \\
\hline
Source only &  0.0856  \\
\hline
TQS \cite{fu2021transferable} & 0.0122 \\
SDM \cite{xie2022learning}& 0.0041  \\
STGADA (Ours) &  0.0044 \\
\bottomrule
\end{tabular}}
\vspace{-3mm}
\end{table}

Per-class classification accuracy results are reported in \cref{table:results}. The results show that our proposed method outperforms the state-of-the-art models. According to per-class accuracy, our method achieves the best performances on the classes \textit{bicycle} and \textit{car}. Although S$^3$VAADA \cite{rangwani2021s3vaada} performs well for the \textit{person} class, the accuracy of that model on the other classes shows that the model overfits to \textit{person} class, which is an abundant class in the source domain dataset. On the other hand, our approach achieves balanced performance for all classes despite the fact that the source dataset is imbalanced. Our method also increases the performance of the base method SDM \cite{xie2022learning} in the \textit{bicycle} class, which has very few samples in the source dataset. It is important to underline this situation because real-world applications usually include imbalanced classes \cite{buda2018systematic, classimbalance}. 

\cref{table:ece} shows the Expected Calibration Error (ECE) \cite{ece} results. If a classifier's observed accuracy matches with its mean confidence, the classifier is considered to be calibrated \cite{ece}.  From the results, we can conclude that the source only model has the highest ECE, which states that the source only model is not calibrated well.  In terms of calibration, our model STGADA and SDM \cite{xie2022learning} outperforms TQS \cite{fu2021transferable} by obtaining much less ECE. In parallel to ECE results, \cref{fig:reliability} demonstrates how far the source only model is from the perfect calibration line. In other words, the source only model cannot achieve the expected accuracy concerning obtained confidence scores. TQS \cite{fu2021transferable} also exhibits unwanted peaks and gaps stating that the model is not calibrated enough. On the other hand, our proposed model STGADA and SDM \cite{xie2022learning} manage to achieve proportional accuracy to confidence scores that the gaps in the reliability diagrams are smaller than those of the source only model and TQS \cite{fu2021transferable}. Since the calibration of the SDM \cite{xie2022learning} is better than that of the TQS \cite{fu2021transferable}, we have chosen the SDM \cite{xie2022learning} as the base model for our approach. Furthermore, the fact that TQS \cite{fu2021transferable} has a high level of complexity with multiple classifiers and a discriminator underlines our base method decision of SDM \cite{xie2022learning}. Consequently, our model is more calibrated than TQS \cite{fu2021transferable} and comparable with SDM \cite{xie2022learning} in addition to achieving higher accuracy. 

\textbf{Qualitative Analysis} We present the visualization of the feature representation on the target thermal IR domain with t-SNE \cite{tsne} for qualitative analysis in \cref{fig:tsne}. The features originating from the source only model cannot be discriminated, while our base method SDM \cite{xie2022learning} and TQS \cite{fu2021transferable} discriminate the samples from different classes. On the other hand, our proposed model STGADA can isolate the samples from the \textit{bicycle} better than those models. The improvement in the discrimination of the samples from \textit{bicycle} reflects itself in the fact that our model achieved the best accuracy in \textit{bicycle} class. Furthermore, we visualized one selected sample from each class, \textit{bicycle}, \textit{car}, and \textit{person} in \cref{fig:tsne}. Our proposed approach tends to select the most uncertain samples that rely upon near the decision boundaries. As we can see from the \cref{fig:tsne}, samples, which are close to different categorical clusters are selected. The visualized examples also confirm this criteria. We can see that the samples from the classes \textit{bicycle} and \textit{car} contain a person object, and the sample from the class \textit{person} contains a bicycle object. This situation makes those samples ambiguous for the trained network, hence, the query function selects those samples.  
\vspace{-3mm}
\section{Conclusion}
\vspace{-3mm}
In this study, we propose a spectral transfer guided active domain adaptation method in order to examine the efficiency of combining visible spectrum and thermal imagery modalities by taking advantage of active domain adaptation. We perform spectral transfer to the source samples with the target samples for additional alignment of two domains. The low frequency component of source samples are exchanged with those of target samples. This extra alignment leads our model to achieve improved performance since the domain shift in RGB-to-thermal task is considerably large. To present our results, we employed the large scale RGB dataset MS-COCO as the source domain and thermal dataset FLIR ADAS as the target domain. Quantitative and qualitative analyses demonstrate that our proposed approach performs better than state-of-the-art active domain adaptation methods by reducing the domain gap between RGB and thermal domains.

{\small
\bibliographystyle{ieee_fullname}
\bibliography{egbib}

\begin{thebibliography}{10}\itemsep=-1pt

\bibitem{agarwal2020contextual}
Sharat Agarwal, Himanshu Arora, Saket Anand, and Chetan Arora.
\newblock Contextual diversity for active learning.
\newblock In {\em ECCV}, 2020.

\bibitem{aghdam2019active}
Hamed~H Aghdam, Abel Gonzalez-Garcia, Joost van~de Weijer, and Antonio~M
  L{\'o}pez.
\newblock Active learning for deep detection neural networks.
\newblock In {\em ICCV}, 2019.

\bibitem{sgada2021}
Ibrahim~Batuhan Akkaya, Fazil Altinel, and Ugur Halici.
\newblock Self-training guided adversarial domain adaptation for thermal
  imagery.
\newblock In {\em CVPRW}, 2021.

\bibitem{balcan2007margin}
Maria-Florina Balcan, Andrei Broder, and Tong Zhang.
\newblock Margin based active learning.
\newblock In {\em COLT}, 2007.

\bibitem{bousmalis2016domain}
Konstantinos Bousmalis, George Trigeorgis, Nathan Silberman, Dilip Krishnan,
  and Dumitru Erhan.
\newblock Domain separation networks.
\newblock {\em NeurIPS}, 2016.

\bibitem{buda2018systematic}
Mateusz Buda, Atsuto Maki, and Maciej~A Mazurowski.
\newblock A systematic study of the class imbalance problem in convolutional
  neural networks.
\newblock {\em Neural Networks}, 2018.

\bibitem{Cho2021MCDALMC}
Jae-Won Cho, Dong-Jin Kim, Yunjae Jung, and In-So Kweon.
\newblock Mcdal: Maximum classifier discrepancy for active learning.
\newblock In {\em IEEE transactions on neural networks and learning systems},
  2021.

\bibitem{das2018unsupervised}
Debasmit Das and CS~George Lee.
\newblock Unsupervised domain adaptation using regularized hyper-graph
  matching.
\newblock In {\em ICIP}, 2018.

\bibitem{imageNet}
Jia Deng, Wei Dong, Richard Socher, Li-Jia Li, Kai Li, and Li Fei-Fei.
\newblock Imagenet: A large-scale hierarchical image database.
\newblock In {\em CVPR}, 2009.

\bibitem{devaguptapu2019borrow}
Chaitanya Devaguptapu, Ninad Akolekar, Manuj M~Sharma, and Vineeth
  N~Balasubramanian.
\newblock Borrow from anywhere: Pseudo multi-modal object detection in thermal
  imagery.
\newblock In {\em CVPR}, 2019.

\bibitem{pascalvoc}
Mark Everingham, Luc Gool, Christopher~K. Williams, John Winn, and Andrew
  Zisserman.
\newblock The pascal visual object classes (voc) challenge.
\newblock {\em IJCV}, 2010.

\bibitem{fft}
M. Frigo and S.G. Johnson.
\newblock Fftw: an adaptive software architecture for the fft.
\newblock In {\em ICASSP}, 1998.

\bibitem{fu2021transferable}
Bo Fu, Zhangjie Cao, Jianmin Wang, and Mingsheng Long.
\newblock Transferable query selection for active domain adaptation.
\newblock In {\em CVPR}, 2021.

\bibitem{dann}
Yaroslav Ganin and Victor Lempitsky.
\newblock Unsupervised domain adaptation by backpropagation.
\newblock In {\em ICML}, 2015.

\bibitem{gilad2006query}
Ran Gilad-Bachrach, Amir Navot, and Naftali Tishby.
\newblock Query by committee made real.
\newblock In {\em NeurIPS}, 2006.

\bibitem{rcnn}
Ross Girshick, Jeff Donahue, Trevor Darrell, and Jitendra Malik.
\newblock Rich feature hierarchies for accurate object detection and semantic
  segmentation.
\newblock In {\em CVPR}, 2014.

\bibitem{flir}
F.~A. Group.
\newblock Flir thermal dataset for algorithm training.
\newblock \url{https://www.flir.com/oem/adas/adas-dataset-form/}.

\bibitem{guan2019unsupervised}
Dayan Guan, Xing Luo, Yanpeng Cao, Jiangxin Yang, Yanlong Cao, George
  Vosselman, and Michael Ying~Yang.
\newblock Unsupervised domain adaptation for multispectral pedestrian
  detection.
\newblock In {\em CVPRW}, 2019.

\bibitem{guo2019domain}
Tiantong Guo, Cong~Phuoc Huynh, and Mashhour Solh.
\newblock Domain-adaptive pedestrian detection in thermal images.
\newblock In {\em ICIP}, 2019.

\bibitem{resnet}
Kaiming He, Xiangyu Zhang, Shaoqing Ren, and Jian Sun.
\newblock Deep residual learning for image recognition.
\newblock In {\em CVPR}, 2016.

\bibitem{kaist}
Soonmin Hwang, Jaesik Park, Namil Kim, Yukyung Choi, and In So~Kweon.
\newblock Multispectral pedestrian detection: Benchmark dataset and baseline.
\newblock In {\em CVPR}, 2015.

\bibitem{BMVC2016_73}
Shu~Wang Jingjing~Liu, Shaoting~Zhang and Dimitris Metaxas.
\newblock Multispectral deep neural networks for pedestrian detection.
\newblock In {\em BMVC}, 2016.

\bibitem{classimbalance}
Justin~M Johnson and Taghi~M Khoshgoftaar.
\newblock Survey on deep learning with class imbalance.
\newblock {\em Journal of Big Data}, 2019.

\bibitem{MCALforclassication}
Ajay~J. Joshi, Fatih Porikli, and Nikolaos Papanikolopoulos.
\newblock Multi-class active learning for image classification.
\newblock In {\em CVPR}, 2009.

\bibitem{Li_2022}
Zhaowen Li, Xu Zhao, Chaoyang Zhao, Ming Tang, and Jinqiao Wang.
\newblock Transfering low-frequency features for domain adaptation.
\newblock In {\em ICME}, 2022.

\bibitem{mscoco}
Tsung-Yi Lin, Michael Maire, Serge Belongie, James Hays, Pietro Perona, Deva
  Ramanan, Piotr Dollar, and Larry Zitnick.
\newblock Microsoft coco: Common objects in context.
\newblock In {\em ECCV}, 2014.

\bibitem{cdan}
Mingsheng Long, Zhangjie Cao, Jianmin Wang, and Michael~I Jordan.
\newblock Conditional adversarial domain adaptation.
\newblock In {\em NeurIPS}, 2018.

\bibitem{long2013transfer}
Mingsheng Long, Jianmin Wang, Guiguang Ding, Jiaguang Sun, and Philip~S Yu.
\newblock Transfer feature learning with joint distribution adaptation.
\newblock In {\em ICCV}, 2013.

\bibitem{ece}
Mahdi Pakdaman~Naeini, Gregory Cooper, and Milos Hauskrecht.
\newblock Obtaining well calibrated probabilities using bayesian binning.
\newblock {\em AAAI}, 2015.

\bibitem{pytorch}
Adam Paszke, Sam Gross, Francisco Massa, Adam Lerer, James Bradbury, Gregory
  Chanan, Trevor Killeen, Zeming Lin, Natalia Gimelshein, Luca Antiga, Alban
  Desmaison, Andreas Kopf, Edward Yang, Zachary DeVito, Martin Raison, Alykhan
  Tejani, Sasank Chilamkurthy, Benoit Steiner, Lu Fang, Junjie Bai, and Soumith
  Chintala.
\newblock Pytorch: An imperative style, high-performance deep learning library.
\newblock In {\em NeurIPS}, 2019.

\bibitem{prabhu2021active}
Viraj Prabhu, Arjun Chandrasekaran, Kate Saenko, and Judy Hoffman.
\newblock Active domain adaptation via clustering uncertainty-weighted
  embeddings.
\newblock In {\em CVPR}, 2021.

\bibitem{rangwani2021s3vaada}
Harsh Rangwani, Arihant Jain, Sumukh~K Aithal, and R~Venkatesh Babu.
\newblock S3vaada: Submodular subset selection for virtual adversarial active
  domain adaptation.
\newblock In {\em ICCV}, 2021.

\bibitem{yolo}
Joseph Redmon, Santosh Divvala, Ross Girshick, and Ali Farhadi.
\newblock You only look once: Unified, real-time object detection.
\newblock In {\em CVPR}, 2016.

\bibitem{fasterrcnn}
Shaoqing Ren, Kaiming He, Ross Girshick, and Jian Sun.
\newblock Faster r-cnn: Towards real-time object detection with region proposal
  networks.
\newblock In {\em NeurIPS}, 2015.

\bibitem{mcdda}
Kuniaki Saito, Kohei Watanabe, Yoshitaka Ushiku, and Tatsuya Harada.
\newblock Maximum classifier discrepancy for unsupervised domain adaptation.
\newblock In {\em CVPR}, 2018.

\bibitem{saponaro2015material}
Philip Saponaro, Scott Sorensen, Abhishek Kolagunda, and Chandra Kambhamettu.
\newblock Material classification with thermal imagery.
\newblock In {\em CVPR}, 2015.

\bibitem{ALforCNN}
Ozan Sener and Silvio Savarese.
\newblock Active learning for convolutional neural networks: A core-set
  approach.
\newblock In {\em ICLR}, 2018.

\bibitem{settles2011theories}
Burr Settles.
\newblock From theories to queries: Active learning in practice.
\newblock In {\em AISTATS}, 2011.

\bibitem{sinha2019variational}
Samarth Sinha, Sayna Ebrahimi, and Trevor Darrell.
\newblock Variational adversarial active learning.
\newblock In {\em CVPR}, 2019.

\bibitem{su2020active}
Jong-Chyi Su, Yi-Hsuan Tsai, Kihyuk Sohn, Buyu Liu, Subhransu Maji, and
  Manmohan Chandraker.
\newblock Active adversarial domain adaptation.
\newblock In {\em WACV}, 2020.

\bibitem{adda}
Eric Tzeng, Judy Hoffman, Kate Saenko, and Trevor Darrell.
\newblock Adversarial discriminative domain adaptation.
\newblock In {\em CVPR}, 2017.

\bibitem{Uzun_2022_CVPR}
Engin Uzun, Ahmet~An{\i}l Dursun, and Erdem Akag\"und\"uz.
\newblock Augmentation of atmospheric turbulence effects on thermal adapted
  object detection models.
\newblock In {\em CVPRW}, 2022.

\bibitem{tsne}
Laurens van~der Maaten and Geoffrey Hinton.
\newblock Visualizing data using t-sne.
\newblock {\em Journal of Machine Learning Research}, 2008.

\bibitem{wang2014new}
Dan Wang and Yi Shang.
\newblock A new active labeling method for deep learning.
\newblock In {\em IJCNN}, 2014.

\bibitem{xie2022active}
Binhui Xie, Longhui Yuan, Shuang Li, Chi~Harold Liu, Xinjing Cheng, and Guoren
  Wang.
\newblock Active learning for domain adaptation: An energy-based approach.
\newblock In {\em AAAI}, 2022.

\bibitem{xie2022learning}
Ming Xie, Yuxi Li, Yabiao Wang, Zekun Luo, Zhenye Gan, Zhongyi Sun, Mingmin
  Chi, Chengjie Wang, and Pei Wang.
\newblock Learning distinctive margin toward active domain adaptation.
\newblock In {\em CVPR}, 2022.

\bibitem{fda}
Y. Yang and S. Soatto.
\newblock Fda: Fourier domain adaptation for semantic segmentation.
\newblock In {\em CVPR}, 2020.

\bibitem{yoo2019learning}
Donggeun Yoo and In~So Kweon.
\newblock Learning loss for active learning.
\newblock In {\em CVPR}, 2019.

\end{thebibliography}
}

\end{document}